\pdfoutput=1
\ifdefined\pdfminorversion\pdfminorversion=7\fi
\documentclass[letterpaper, 10pt, conference]{ieeeconf}

\IEEEoverridecommandlockouts  
\useRomanappendicesfalse      

\makeatletter
\def\@IEEEauthorblockconfadjspace{-1.7em}
\makeatother

\makeatletter
\def\@IEEEsectpunct{\ }
\makeatother

\usepackage[dvipsnames,table]{xcolor}
\usepackage{amsmath}
\usepackage{amssymb}
\usepackage{graphicx}
\usepackage{float}

\usepackage{booktabs}
\usepackage{soul}
\usepackage{pifont}
\usepackage{adjustbox}

\usepackage{wrapfig}
\usepackage{tcolorbox}
\usepackage{tabularx}

\newcommand{\tna}{\textcolor{black!40}{--}}
\usepackage{siunitx}
\definecolor{qcol}{HTML}{003049}
\newcommand{\Q}[1]{\textcolor{qcol}{\textbf{#1}}}

\makeatletter
\def\caption@documentclass{standard}
\makeatother

\usepackage{wrapfig}

\usepackage{graphicx}
\usepackage{amsmath}
\usepackage{amssymb}
\usepackage{bm}
\usepackage{mathtools}

\def\namedlabel#1#2{\begingroup
    #2%
    \def\@currentlabel{#2}%
    \phantomsection\label{#1}\endgroup
}
\makeatletter
\newcommand{\leqnomode}{\tagsleft@true\let\veqno\@@leqno}
\makeatother
\newcommand\PMPODE[1]{\hyperref[PMPODE]{$\textbf{ODE}_{#1}$}\xspace}

\newcommand\PMPODEg[1]{\hyperref[PMPODEg]{$\textbf{ODE}^g_{#1}$}\xspace}
\newcommand\OCPg[1]{\hyperref[OCPg]{$\textbf{OCP}^g_{#1}$}\xspace}

\newcommand\OCP[1]{\hyperref[OCP]{$\textbf{OCP}_{#1}$}\xspace}
\newcommand\BVP[1]{\hyperref[BVP]{$\textbf{BVP}_{#1}$}\xspace}

\usepackage{tabularx}
\newcolumntype{C}{>{\centering\arraybackslash}p{19mm}}
\newcolumntype{G}{>{\centering\arraybackslash}p{4mm}}

\usepackage[font=small,labelfont=bf]{caption}

\usepackage{xspace}

\let\labelindent\relax
\usepackage{enumitem} %

\usepackage{amssymb}%
\usepackage{algorithm}
\makeatletter
\renewcommand*{\ALG@name}{Alg.}
\makeatother
\usepackage[noend]{algpseudocode}

\usepackage{xpatch} %

\newcommand\mydots{\hbox to 1em{.\hss.\hss.}}

\providecommand{\Q}{\mathbf{Q}} 

\newtheorem{preremark3}{Theorem}[section]

\usepackage{mdframed}
\usepackage{lipsum}
\newmdtheoremenv{theo}{Theorem}

\usepackage{multirow}

\usepackage[stretch=30,shrink=30]{microtype}

\begingroup\fontencoding{OT1}\fontfamily{ptm}\selectfont\endgroup
\DeclareFontShape{OT1}{ptm}{m}{scit}{<->ssub*ptm/m/sc}{}

\makeatletter\let\NAT@parse\undefined\makeatother 
\usepackage[square,numbers,sort&compress]{natbib}
\usepackage{url}
\usepackage{hyperref}
\definecolor{mydarkblue}{rgb}{0,0.08,0.45}
\hypersetup{
  pdfborder=0 0 0,
  pdfpagemode=UseNone,
  colorlinks=true,
  linkcolor=mydarkblue,
  citecolor=mydarkblue,
  filecolor=mydarkblue,
  urlcolor=mydarkblue,
  pdfview=FitH}
\hypersetup{pdfauthor={Finley R. Holt, Luis A. Pabon, John Irvin Alora, Jonas Frey, Marco Pavone},pdftitle={Structured World-State Reasoning for Agentic Robotic Search},pdfsubject={},pdfkeywords={}}
\usepackage[capitalize]{cleveref} 
\crefname{figure}{Fig.}{Figs.}   
\Crefname{figure}{Fig.}{Figs.}
\definecolor{corlhighlight}{RGB}{255,235,120}

\definecolor{barslate}{HTML}{2C6E8F}

\title{\LARGE \bf
Structured World-State Reasoning for\\Agentic Robotic Search
}

\author{Finley R.~Holt$^{1}$, Luis A.~Pabon$^{1}$, John Irvin Alora$^{1}$, Jonas Frey$^{1}$, and Marco Pavone$^{1}$
\thanks{\raggedright $^{1}$Department of Aeronautics and Astronautics, Stanford University
        (e-mail: {\tt\small \{frholt, lpabon, jjalora, jonfrey, pavone\}@stanford.edu}).}%
}

\IEEEaftertitletext{%
  \vspace{-1.0\baselineskip}%
  {\centering
    \captionsetup{type=figure}%
    \includegraphics[width=0.94\textwidth]{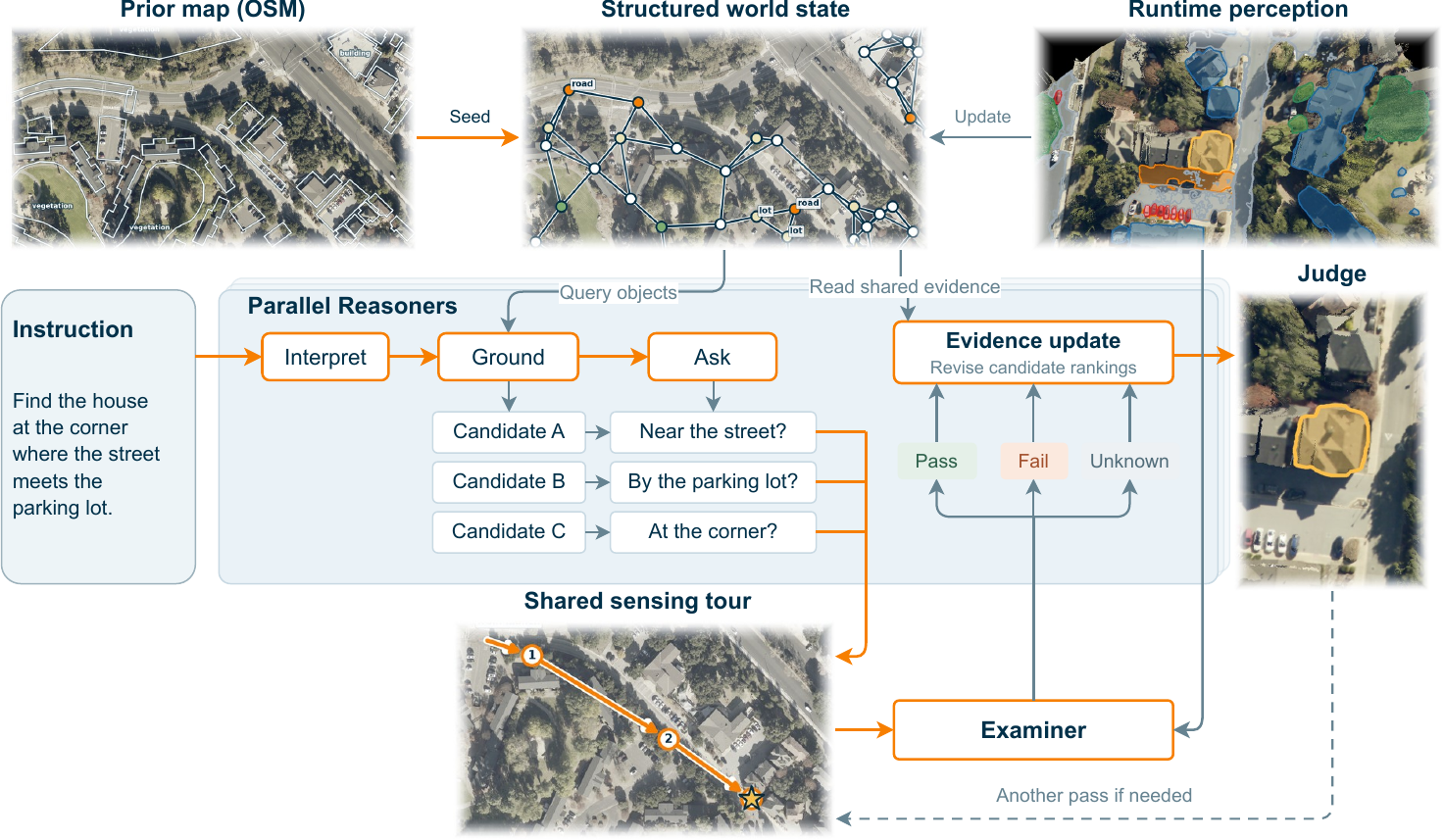}\endgraf 
    \caption{\textbf{WORLDS overview.} Reasoners ground an instruction in a Structured World State (SWS), propose competing candidates and evidence requests, and the drone executes a sensing tour. If anything of interest is found, perception can update the graph. After the tour, an Examiner evaluates the observations and the Reasoners revise their rankings. Finally, a Judge selects the target or requests more evidence.}%
    \label{fig:herofig}%
  }%
  \vspace{0.2\baselineskip}%
}

\begin{document}
\maketitle
\thispagestyle{empty}
\pagestyle{empty}


\begin{abstract}
Long-horizon robotic search must resolve natural language against heterogeneous, incomplete, and often ambiguous evidence: textual information, prior maps, and observations arriving over time. The core challenge is to contextualize these streams and decide where to gather evidence before selecting a target. We present WORLDS: World-state Observation and Reasoning for Language-guided Discovery and Search, a framework that grounds reasoning in a persistent graph initialized from geospatial priors and updated by perception. Parallel Reasoners maintain competing candidate interpretations and request evidence to distinguish between them. We collect and process the requested observations with a multimodal Examiner, after which a Judge selects a grounded target or requests another pass. WORLDS achieves $51.8\%$ navigation success across all $5{,}311$ CityNav test episodes, the highest reported success rate, exceeding the previous published best by $15.7$ percentage points under an OSM-only, high-resolution orthographic protocol. On $1{,}000$ shared episodes, it achieves $50.0\%$ versus $27.9\%$ for the strongest adapted baseline using the same model, prior, sensing stack, and movement budget. Observation-based verification by the Examiner contributes $5.9$ points of this success, and at a reduced reasoning-effort setting WORLDS still exceeds the adapted GeoNav baseline by $18.8$ points while generating fewer tokens. We also demonstrate WORLDS on a quadrotor, which flies the generated sensing waypoints and grounds three language targets, including a vehicle absent from the map, from its onboard imagery.
\end{abstract}

\section{Introduction}
\label{sec:intro}

``\textit{Find my red SUV. It's parked on the top deck of the garage.}'' A person hearing this builds a \textit{mental model} at once: the garage is on the map, the vehicle is not, and the way to find it is to go to the garage and look. Large Language Models (LLMs) give robots the world knowledge and common-sense reasoning~\cite{ahn2022can,driess2023palm} to interpret such an instruction; the open challenge is to act on information that points to places and objects the robot has not yet seen. In search and rescue, that evidence is incomplete, scattered, and arrives over time across multiple \textit{streams}: textual information, geospatial priors, and onboard perception. For search, choosing an initial grounding also creates an evidence-gathering problem: the named anchor may match several map features, the vehicle may be one of several red cars in view, and the robot must preserve plausible alternatives and visit locations that distinguish them. \Cref{fig:realworld} shows exactly this instruction resolved from a physical quadrotor's imagery.

To this end, we introduce \textbf{WORLDS}: \textbf{W}orld-state \textbf{O}b\-ser\-va\-tion and \textbf{R}ea\-son\-ing for \textbf{L}an\-guage-guided \textbf{D}is\-cov\-ery and \textbf{S}earch, a framework built on three principles. An ambiguous instruction admits several groundings, so WORLDS keeps them alive as competing hypotheses rather than committing early. What separates the hypotheses are predicates that only an observation can settle, so each hypothesis states the view it needs, and because observations cost flight time, those requests are merged into one shared sensing tour. Spatial facts are kept apart from judgments: a persistent \emph{Structured World State} (SWS), seeded from the map and written at runtime only by perception, holds what is known, while the agents that interpret, verify, and adjudicate read from it and revise their rankings against the same evidence.

The core contributions of this paper are:
\begin{enumerate}
    \item \textbf{A world-state architecture for language-guided search.} WORLDS maintains a persistent Structured World State seeded from OpenStreetMap and written at runtime only by perception. Parallel Reasoners hold competing groundings of the instruction, attach predicate questions to graph objects, and share one sensing tour; a multimodal Examiner grades each observation, and a Judge commits or requests more evidence, so a target absent from the map is admitted once it is observed.
    \item \textbf{State-of-the-art CityNav results under a controlled comparison.} WORLDS reaches $51.8\%$ success on all $5{,}311$ test episodes, $15.7$ points above the best published result under an OSM-only orthographic protocol, and $50.0\%$ against $27.9\%$ for the stronger of two adapted baselines run with the same model, prior, observation backend, and movement budget on $1{,}000$ shared episodes.
    \item \textbf{An empirical account of why the design works.} Paired ablations attribute $5.9$ points to observation-based verification and show that more Reasoners request $2.5\times$ as many map objects and raise candidate coverage by 23 points at $0.26$ additional sensing stops; a reasoning-effort sweep shows the advantage over adapted GeoNav persisting at fewer generated tokens; and $35.4\%$ of successful commitments select objects added by perception. A quadrotor deployment confirms the pipeline on real imagery, including a vehicle absent from the map.
\end{enumerate}

\section{Related Work}
\label{sec:related}

\textbf{Foundation models as policies and planners.}
End-to-end policies map observations and language to actions~\citep{zitkovich2023rt,kim2024openvla,cai2025flightgpt}. Foundation-model planners select skills using affordances~\citep{ahn2022can}, incorporate language feedback~\citep{huang2022inner}, or ground plans in images and robot state~\citep{driess2023palm}. Methods that quantify grounding uncertainty~\citep{ren2023knowno} or maintain competing referents~\citep{shridhar2018ingress} can resolve ambiguity through interaction with a human. WORLDS addresses the complementary problem of retaining alternative groundings while gathering physical observations that distinguish them.

\textbf{Structured spatial memory and uncertain language.}
Object-centric maps provide persistent spatial context~\citep{huang2023vlmaps,chen2023nlmap}; ConceptGraphs~\citep{gu2024conceptgraphs} builds open-vocabulary 3D scene graphs, and SayPlan~\citep{ranasayplan} supports LLM planning over large graphs. SLOOP~\citep{zheng2021sloop} treats ambiguous spatial language as a stochastic observation in a partially observable city-scale search problem, using OpenStreetMap and online planning. WORLDS also starts from a geospatial prior, but represents uncertainty through language-generated candidate bindings and evidence requests rather than a calibrated probabilistic belief. Its division of responsibilities is related to blackboard architectures~\citep{hayesroth1985blackboard}: agents share spatial context, with runtime object writes restricted to the perception pipeline.

\textbf{Active perception and agentic verification.}
Active perception~\citep{bajcsy1988active} and next-best-view planning~\citep{bircher2016receding} choose observations to support a task, including object search~\citep{aydemir2013active}, semantic exploration~\citep{chaplot2020object}, and embodied question answering~\citep{das2018embodied}. Language-guided agents also connect map goals to observations without task-specific training~\citep{shah2022lmnav}. Interleaved reasoning and action~\citep{yao2022react} and external checking of model outputs~\citep{lifshitzmulti} motivate an explicit verification step. WORLDS combines these ideas by generating task-dependent predicates and segmentation concepts, sharing sensing coverage across Reasoners, and carrying candidate-specific evidence between stops.

\textbf{Aerial vision-language navigation.}
AerialVLN~\citep{liu2023aerialvln}, AVDN~\citep{fan2023avdn}, and CityNav~\citep{lee2025citynav} support language-conditioned flight. Learned approaches include FlightGPT~\citep{cai2025flightgpt} and HTNav~\citep{fan2026htnav}. Among training-free methods, STMR~\citep{stmr} uses semantic--topo--metric maps; GeoNav~\citep{geonav} maintains evolving spatial memory through a global cognitive map and local scene graphs; and ViSA~\citep{visa} separates perception, visual verification, and execution with structured image prompts. STMR, AirHunt~\citep{airhunt}, and RAVEN~\citep{kim2026raven} also investigate physical deployment. WORLDS differs in how hypotheses and observations interact: each Reasoner's competing candidate bindings generate evidence requests tied to graph objects, one shared tour answers them, and the returned observations update every Reasoner's ranking through the persistent graph before the Judge commits. GeoNav commits to a landmark region before visual search; ViSA carries only a language guidance signal between rounds; and CLOSER-VLN~\citep{closervln} verifies proposed actions and revises them by retrieval, whereas WORLDS verifies competing object interpretations against requested, shared physical observations. We adapt GeoNav and ViSA as shared-input baselines in \Cref{sec:exp-citynav}.

\section{Method}
\label{sec:framework}

\subsection{Problem Statement}
\label{sec:method_problem}

We cast language-guided search as target identification followed by navigation. Given an instruction $q$, a robot pose $\xi$, a prior map, observations, and a sensing/navigation contract $\kappa$ (specifying the observation geometry, allowed motion, and map-to-world transform), WORLDS gathers evidence and selects a graph object $o^\star$ with a world position, or returns no commitment. The objective is to identify the intended object and reach it within an action budget.

\subsection{Structured World State}
\label{sec:method_sws}

WORLDS combines three sources of evidence: the instruction, a prior map, and observations acquired during flight. Competing interpretations of the instruction can only be compared if all agents reason over the same evidence, so we maintain it in a single shared representation. The SWS is a persistent graph $\mathcal{G}=(\mathcal{O},\mathcal{E})$ whose nodes $\mathcal{O}$ represent objects and whose edges $\mathcal{E}$ encode spatial relations between them. Two design decisions follow from this role. First, the graph stores facts rather than judgments. Thus, it is initialized from the map and updated only by perception, so no reasoning agent can alter the evidence available to the others. Second, prior and observed objects are stored as distinct nodes, which preserves the provenance of every fact and allows a target missing from the map to be added once it is observed.

\paragraph{Representation and initialization.}
Each node stores a stable identifier, semantic class, names and tags, footprint, centroid, and an evidence summary comprising observation count, confidence, and references to observations. We initialize the graph from OpenStreetMap (OSM) geometry and tags, which allows the Reasoners to ground anchors and enumerate candidates before any observation is acquired.

\paragraph{Spatial relations.}
Instructions locate targets by relation: \emph{west of the park}, \emph{next to the road}, \emph{inside the campus}. The graph therefore precomputes three footprint relations over prior objects, \emph{contains}, \emph{overlaps}, and \emph{adjacent}, and answers metric queries by centroid distance and directional queries by eight compass bearings. The Reasoners' retrieval tools filter candidates with these relations, and the same geometry supplies the deterministic checks that accompany the Examiner's visual verdicts at each stop.

\paragraph{Updates from perception.}
At each sensing stop, an open-vocabulary segmenter is prompted with the concepts proposed by the Reasoners, and the resulting masks are projected into the map frame, so that detections enter the graph in the coordinate frame of the prior. Repeated detections of the same object across stops, matched by class and position, are merged into a single runtime node whose footprint and evidence summary are refined with each observation; unmatched detections instantiate new nodes. Detections are never merged into prior features.

\paragraph{Agentic retrieval.}
A city-scale prior contains thousands of objects, of which few are relevant to a given hypothesis, so agents access the graph through two retrieval tools. $\textsc{Find}$ returns the identifiers of objects matching a description, filtered by name, tag, class, provenance (prior or runtime), size, and spatial relation to one or more anchors, the reference objects named in the instruction; for \emph{the warehouse west of the park}, a Reasoner first retrieves the park and then queries for warehouses west of it. $\textsc{Inspect}$ returns the stored information for one identifier, including attributes, neighboring objects with distances and bearings, relations, and observation context, which the Reasoners use to compare candidates. Further tools let Reasoners record their evidence and rankings in per-agent records, and every identifier they cite is validated against the graph. We do not let them modify shared object facts, to prevent any single hypothesis from biasing the evidence available to the others.

\subsection{Agentic Workflow}
\label{sec:method_workflow}

An ambiguous instruction admits several groundings, and the evidence that separates them must be gathered by flight. The workflow therefore proceeds in three stages (\Cref{fig:herofig}, Algorithm~\ref{alg:worlds}). Reasoners interpret the instruction (\textsc{Interpret}), ground it in the graph as competing candidates (\textsc{Ground}), and state the observations that would distinguish them (\textsc{Ask}); one shared sensing tour acquires those observations; and an Examiner and a Judge turn them into a verified target. We launch $L$ Reasoners in parallel, differing only in sampling seed, and retain the first $M$ whose plans contain at least one observation request that names a graph object; independent sampling diversifies the hypotheses. If no plan is valid, the episode aborts before flight.

\begin{algorithm}[t]
\caption{WORLDS decision sequence and action scoring}
\label{alg:worlds}
\begin{algorithmic}[1]
\Require instruction $q$, graph $\mathcal{G}$, pose $\xi$, contract $\kappa$, budget $B$, launch/retain counts $(L,M)$
\For{each of $L$ Reasoners \textbf{in parallel}}
  \State \textsc{Interpret} $q$; \textsc{Ground} with \textsc{Find}/\textsc{Inspect}
  \State \textsc{Ask}: attach predicate questions to graph objects
\EndFor
\State freeze concept union; retain first $M$ valid plans
\State $T \gets$ shared geometric coverage and tour ordering
\Loop
  \For{each stop in $T$, in order}
    \State acquire observation; update $\mathcal{G}$ from perception
    \State Examiner grades each observed object
    \State Reasoners update evidence and rankings
  \EndFor
  \State Judge: select target or request another pass
  \If{target committed or no extension is permitted}
    \State \textbf{break}
  \EndIf
  \State $T \gets$ appended centroid stops; count one extension
\EndLoop
\State validate target; apply fallback if needed
\State \textbf{Score:} replay all recorded sensing stops under $\kappa$
\State append commitment leg if committed before motion ended
\State terminate at budget $B$; score reached pose
\end{algorithmic}
\end{algorithm}

\paragraph{Interpretation (\textsc{Interpret}).}
Each Reasoner reads $q$ and any accompanying text and identifies what is sought, the landmarks that locate it, which we call \emph{anchors}, and the features that distinguish it; for the instruction in \Cref{sec:intro}, these are the red SUV, the garage, and its top deck. It keeps several readings where the language is ambiguous. It also proposes short segmentation concepts, i.e., text prompts for the open-vocabulary segmenter such as \emph{red SUV}. Their union over all $L$ Reasoners, capped at 32, is frozen for the episode, so perception searches at every stop for exactly the objects the hypotheses refer to.

\paragraph{Map grounding (\textsc{Ground}).}
Each Reasoner resolves its anchors to graph objects and enumerates candidate targets with \textsc{Find} and \textsc{Inspect}, recording a ranked list of candidates, a justification for each, and any phrase it could not resolve. A target absent from the prior cannot be enumerated; the Reasoner then grounds the surrounding context instead, which directs the sensing, and perception adds the target once it is observed.

\paragraph{Question--viewpoint proposal (\textsc{Ask}).}
The remaining candidates differ in properties that the map cannot settle. Each Reasoner therefore states these properties as yes/no questions about specific graph objects, e.g., \emph{Is this a warehouse?} or \emph{Does it border the river road?} Naming the object makes the request concrete: it fixes the viewpoint, a top-down (orthographic), north-up view centered on the object at a bounded altitude, and it yields a question the Examiner can check. Spatial questions such as adjacency are also answered by geometric computation on the graph, independently of the vision model.

\paragraph{Tour synthesis.}
Observations cost flight time, so the requests of all retained Reasoners are served by one tour. We choose the fewest views that together cover all requests, a set-cover problem, and order them into a short open path. The tour is fixed during a pass, one traversal of its stops; the Judge can add stops between passes.

\paragraph{Execution and verification.}
At each stop, perception updates the SWS and the open questions are graded. We assign grading to a single Examiner rather than to the Reasoners, so that every hypothesis is judged against the same observation by the same standard. For each observed object, the Examiner answers all questions about that object in one round. Its inputs are the instruction, the questions, the graph context, its memory from earlier stops, and up to two views in which the object's map footprint and the detected masks are drawn~\citep{yang2023setofmark}. It returns \textsc{Pass}, \textsc{Fail}, or \textsc{Unknown} with a confidence and, if a different object satisfies the question, that object's identifier. Each Reasoner then updates its leading candidate, alternatives, and supporting evidence, treating the verdicts as weighted evidence rather than as decisions.

\paragraph{Adjudication (Judge).}
The Reasoners that propose candidates do not decide between them; a single Judge does, after each pass. It reads the Reasoners' latest records, all Examiner verdicts, and the graph context around the objects the Examiner named, and ranks the candidates on the observed evidence. It may select any object in the graph with a world position, whether from the map or from perception, including one absent from every initial ranking, so evidence can override the Reasoners' initial vote (\Cref{fig:trace}). If the evidence is insufficient, the Judge requests further observations instead; up to two such extensions add stops over the requested objects, and the Reasoners keep their context. If the Judge returns no valid identifier, it is re-run under tighter constraints, then the candidate the Reasoners agree on most is taken, then the best-ranked candidate before flight. An episode without a groundable target is recorded as no commitment.

\section{Experiments}
\label{sec:exp-bench}

We evaluate WORLDS on CityNav and report a field case study on a physical quadrotor. All runs executed for this study use the same evaluation implementation.

\begin{figure}[b]
\centering
\includegraphics[width=\linewidth]{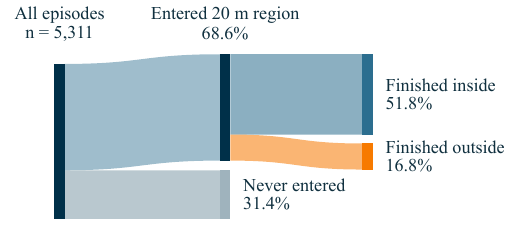}
\caption{\textbf{Trajectory outcomes at 200 actions.} Entry and final position are measured against the target's 20\,m region. Shares and widths derive from rounded aggregate rates over all $5{,}311$ episodes.}
\label{fig:failures}
\end{figure}

\begin{figure*}[t]
\centering
\includegraphics[width=\textwidth]{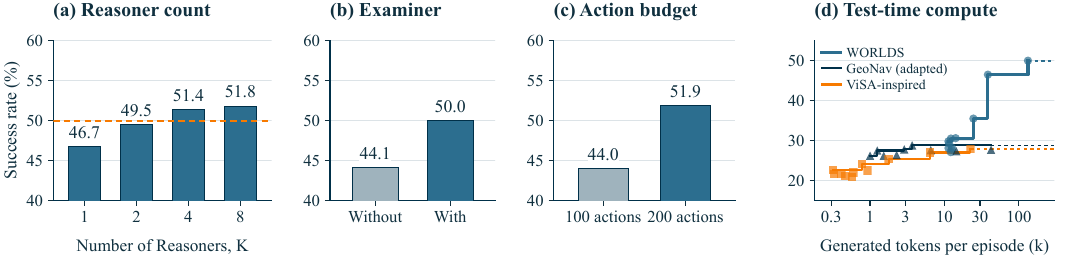}
\caption{\textbf{Ablations, budget sensitivity, and test-time compute.} Vertical axes span $40$--$60\%$ in (a--c) and $15$--$55\%$ in (d). \textbf{(a,b)} The same $1{,}000$ episodes at 200 actions. The control (orange dashed line in a) launches five Reasoners and retains the first three; the $K$ arms retain all $K$. The Examiner adds 5.9 points. \textbf{(c)} Both budgets use the same $5{,}300$ replayable episodes, excluding 11 preflight aborts; the 200-action replay gains 7.9 points. \textbf{(d)} The same $1{,}000$ episodes at 200 actions. Each harness is frozen and its reasoning-effort hyperparameter is varied from 0 to 0.99, which sets how many generated tokens (including thinking) the model spends; every point is one setting. The line is the best success rate the harness reaches within a per-episode token budget, dashed past the largest spend measured. WORLDS keeps converting tokens into success from 11k to 130k tokens per episode; ViSA stops gaining at 20k, and GeoNav gains nothing across a $40\times$ range.}
\label{fig:compute}
\end{figure*}

\subsection{Benchmark performance on CityNav}
\label{sec:exp-citynav}

\paragraph{Inputs and movement contract.}
CityNav~\citep{lee2025citynav} provides language-conditioned aerial search episodes over SensatUrban~\citep{sensaturban} reconstructions of Birmingham and Cambridge, with object descriptions from CityRefer~\citep{cityrefer}. We evaluate all $5{,}311$ \texttt{test\_unseen} episodes using an OSM-only prior~\citep{haklay2008openstreetmap}: the graph contains OSM geometry and tags, rasterized at 2\,m. Beyond the task instruction, no CityRefer object catalog, per-object descriptions, or answer-key coordinates enter the graph or tools. Observations are north-up orthographic crops from a 0.1\,m/pixel raster, capped at 2{,}048 pixels and resampled to $2{,}016\!\times\!2{,}016$. Their square footprint has width twice the height above the local surface, corresponding to a nominal $90^\circ$ footprint. This high-resolution orthographic input setting differs from native published baseline protocols. The shared action interface comprises forward 5\,m, turns of $\pm30^\circ$, ascent/descent of 2\,m, and stop; the primary budget is 200 actions.

\paragraph{Trajectory scoring.}
We score recorded decisions with a discrete executor that replays every sensing stop in its original order, followed by a final commitment leg when a target was committed before motion ended; a later fallback selection leaves the scored pose unchanged. Each leg starts from the pose the executor reached on the previous one. Movement to acquire observations and movement toward the committed target both consume the action budget; if the budget expires during sensing or final navigation, the pose actually reached is scored. The executor's pose never feeds back to the model; images are rendered at the requested waypoints, not at the reached poses. Completed sensing legs end a median 1.44\,m and at most 2.59\,m from the requested centers, with altitude error below 1\,m. The 100-action budget study replays the same recorded observations, tours, and commitments without rerunning any model, so it evaluates the recorded decisions under a shorter movement allowance rather than a policy replanned for it.

We report final-pose success rate (SR), oracle success rate (OSR: any pose along the trajectory enters the target region), and success weighted by path length (SPL)~\citep{anderson_spl}. The success region is a horizontal 20\,m radius around the annotated target. SPL uses planar traveled distance and the direct start-to-target distance. These are positional metrics, not object-identity accuracy. All 11 full-split no-commitment episodes abort before takeoff and remain in the denominator at their unchanged start pose.

\paragraph{Implementation.}
WORLDS's agent roles use the frozen Inkling-Small-NVFP4 model~\citep{inkling_small}, a multimodal mixture-of-experts model with 276B total and 12B active parameters, served with SGLang at a reasoning effort of 0.99. The full-split run used six B300 model replicas across two GB300 nodes. Runtime perception uses SAM~3.1~\citep{sam31} in image mode, with $1{,}008$-pixel tiles, 20\% overlap, and 0.5 nonmaximum-suppression threshold. No weights are trained on this benchmark. The control launches five Reasoners and retains the first three valid plans; completion-order selection is a latency heuristic. The grounding, per-stop Reasoner, and Examiner tool loops allow up to 60, 6, and 16 turns, respectively. The $K$-sweep launches and retains all $K$ Reasoners; the no-Examiner arm retains the control's five/three configuration; the reasoning-effort sweep keeps the control configuration and varies only the effort setting.

\begin{table}[t]
\centering
\caption{\textbf{CityNav test results.} Published rows are self-reported under each method's own inputs and step budget. WORLDS rows use the OSM-only prior, shared sensing/model stack, and 200-action replay; the lower block uses identical episode IDs and the adapted baselines described in the text. SR and OSR are percentages; SPL is a fraction.}
\label{tab:citynav}
\small
\setlength{\tabcolsep}{3.5pt}
\begin{tabular}{lrrrr}
\toprule
Method & $n$ & SR$\uparrow$ & OSR$\uparrow$ & SPL$\uparrow$ \\
\midrule
\multicolumn{5}{l}{\textit{Published, own inputs and budget}} \\
GeoNav~\citep{geonav} & \tna & 25.9 & 41.6 & .160 \\
CLOSER-VLN~\citep{closervln} & \tna & 32.0 & 43.5 & .213 \\
ViSA~\citep{visa} & \tna & 36.1 & 43.4 & .273 \\
\midrule
\textbf{WORLDS} & 5{,}311 & \textbf{51.8} & \textbf{68.6} & \textbf{.445} \\
\midrule
\multicolumn{5}{l}{\textit{Shared episodes and inputs}} \\
\textbf{WORLDS (control)} & 1{,}000 & \textbf{50.0} & \textbf{65.6} & \textbf{.427} \\
\quad No Examiner & 1{,}000 & 44.1 & 61.8 & .385 \\
\quad One Reasoner ($K=1$) & 1{,}000 & 46.7 & 62.4 & .411 \\
ViSA-inspired (adapted) & 1{,}000 & 27.9 & 56.5 & .212 \\
GeoNav (adapted) & 1{,}000 & 27.7 & 60.7 & .163 \\
\bottomrule
\end{tabular}
\end{table}

\paragraph{Controlled baseline comparison.}
To compare policies rather than input pipelines, we hold the model (Inkling) and its served endpoint (a 262{,}144-token context with up to 40{,}000 output tokens per reasoning request; auxiliary formatting and compaction calls use smaller caps), prior (OSM-only), observation backend and perception model (orthographic rendering and SAM~3.1), action executor, movement budget, and episode IDs fixed and vary only the policy. The two baselines, the ablations, and the control all run on the same $1{,}000$ episode IDs: the $1{,}000$ that follow the first 2{,}656 of a seed-1234 shuffle of the test split. The ViSA-inspired policy follows the published perception--verification--execution organization~\citep{visa}, with SAM masks in place of VLM-proposed detections and OSM landmarks as context. The adapted GeoNav policy starts from the released code~\citep{geonav}, with SAM~3.1 in place of GroundingDINO/MobileSAM, the shared observation resolution, and OSM landmarks. Each policy decides how many model calls it makes, so inference expenditure differs across arms; \Cref{tab:arms} reports it.

\paragraph{Results.}
WORLDS achieves $51.8\%$ SR, $68.6\%$ OSR, and $0.445$ SPL on all $5{,}311$ episodes (\Cref{tab:citynav}). The best published CityNav test results are ViSA at $36.1\%$ SR and $0.273$ SPL~\citep{visa}, CLOSER-VLN at $32.0\%$~\citep{closervln}, and GeoNav at $25.9\%$~\citep{geonav}, each under its own input protocol and step budget. WORLDS exceeds the best published SR by $15.7$ points and the best published SPL by $0.172$; its 100-action replay ($44.0\%$, \Cref{fig:compute}c) also exceeds the published best. On the shared $1{,}000$, the control reaches $50.0\%$ SR, exceeding the ViSA-inspired and GeoNav adaptations by $22.1$ and $22.3$ points. Its OSR is $65.6\%$ versus $56.5\%$ and $60.7\%$, respectively. Among episodes that enter the target region, $76.2\%$ of the control's trajectories finish there, compared with $49.4\%$ and $45.6\%$ for the adaptations.

\paragraph{Examiner ablation.}
Observation-based verification improves budgeted success by $5.9$ percentage points, from $44.1\%$ to $50.0\%$ SR. Of $1{,}000$ paired episodes, 149 succeed only with the Examiner, 90 only without it, and 351 with both. The Examiner also raises OSR from $61.8\%$ to $65.6\%$ and SPL from $0.385$ to $0.427$ (\Cref{tab:citynav}). The no-Examiner arm retains perception, runtime graph updates, geometric checks, persistent Reasoners, and the Judge, with stop and Judge prompts adjusted to the module's absence. Perception-added nodes carry a large share of success: the full run selects them in $1{,}787$ of $5{,}311$ episodes, of which 975 succeed ($54.6\%$ SR), accounting for $35.4\%$ of all successful episodes; prior-origin selections succeed in $1{,}778$ of $3{,}513$ ($50.6\%$). These strata describe selected-node provenance and positional success; a detected node can coincide with an object also represented in the prior.

\begin{table}[t]
\centering
\caption{\textbf{Candidate coverage and shared sensing.} All $1{,}000$ episodes ran in each arm; the table pairs the 922 that produced a plan and a Judge record in all four. Objects and stops are means over the initial tour. Coverage is a lower bound on the fraction with a Reasoner-ranked candidate inside the target's 20\,m region at adjudication; at most 1.3\% of episodes per arm remain unresolved from logged coordinates.}
\label{tab:mechanism}
\small
\setlength{\tabcolsep}{4pt}
\begin{tabular}{rrrr}
\toprule
$K$ & \shortstack{Requested\\map objects} & \shortstack{Sensing\\stops} & \shortstack{Candidate\\coverage (\%)} \\
\midrule
1 & 2.28 & 1.13 & 54.0 \\
2 & 3.27 & 1.20 & 61.1 \\
4 & 4.41 & 1.26 & 69.2 \\
8 & 5.65 & 1.39 & 77.2 \\
\bottomrule
\end{tabular}
\end{table}

\begin{figure}[t]
\begin{tcolorbox}[colback=white,colframe=qcol,boxrule=0.6pt,arc=1pt,left=4pt,right=4pt,top=3pt,bottom=3pt,fonttitle=\bfseries\footnotesize,title={Recorded evidence-to-decision trace (CityNav, eight Reasoners)}]
\footnotesize
\setlength{\parskip}{2pt}
\textbf{Instruction.} ``this is a long and high building with a gray gable roof. The long side of this building has the front court ground, and the short side is the back lawn ground.''

\textbf{Ground.} Eight Reasoners resolve the two named grounds as anchors and split on the building: candidate A receives four initial votes, candidate B one, and three other objects one each.

\textbf{Ask.} The union of requests names six map objects; shared coverage serves them with three initial stops, and the Judge requests two extensions.

\textbf{Examine.} Candidate A: four initial votes $\to$ the long-side/short-side relation to the two grounds is contradicted at two stops (eight failed adjacency predicates) and no roof evidence is found. Candidate B: one initial vote $\to$ its long side is verified adjacent to the front court and a gray roof is observed at runtime; the gable shape remains unresolved.

\textbf{Judge.} With the initial rankings still in its input, the Judge selects candidate B on the observed evidence. Candidate B lies 2.4\,m from the annotated target; candidate A lies 70.8\,m away. The endpoint was reached during sensing, so no separate commitment leg was needed.
\end{tcolorbox}
\caption{\textbf{Evidence overriding the vote leader.} One recorded $K=8$ CityNav episode, selected after inspecting outcomes; candidate A is the four-vote leader and candidate B the selected minority candidate.}
\label{fig:trace}
\end{figure}

\begin{figure*}[t]
\centering
\includegraphics[width=\textwidth]{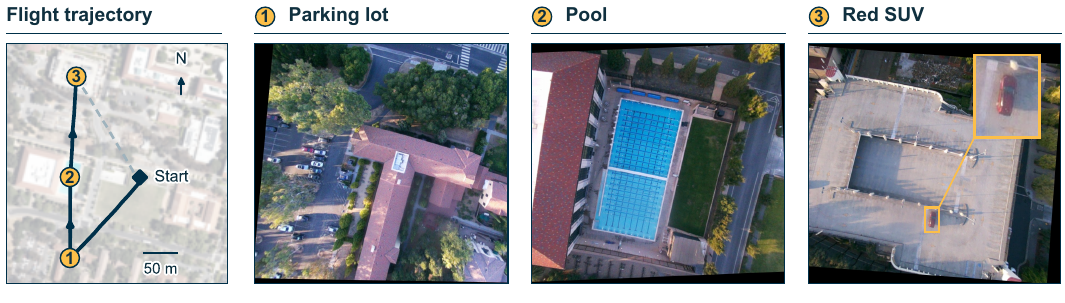}
\caption{\textbf{Grounding from physical aerial observations.} The recorded flight trajectory (left) visits three generated sensing waypoints; solid and dashed segments denote mission execution and pilot-initiated return, over a public-domain NAIP aerial basemap shown for orientation only. Numbered frames show the parking lot, pool, and red SUV selected from onboard imagery; the SUV, absent from the OSM prior, was added by perception and selected by the Judge (gold box, enlarged view). The pool is selected despite a conflicting directional cue.}
\label{fig:realworld}
\end{figure*}

\paragraph{Reasoner count and movement budget.}
Retaining more parallel Reasoners broadens what the system asks to see while the shared planner keeps sensing compact. Across the $K$ arms (\Cref{tab:mechanism}), raising $K$ from one to eight increases the mean number of distinct requested map objects from $2.28$ to $5.65$, while the shared tour grows only from $1.13$ to $1.39$ sensing stops; the fraction of episodes with a Reasoner-ranked candidate inside the target's 20\,m region rises from at least $54.0\%$ to $77.2\%$. Requested objects include anchors as well as candidate targets, and coverage is positional rather than object identity. On all $1{,}000$ episodes, SR rises from $46.7\%$ at $K=1$ to $49.5\%$, $51.4\%$, and $51.8\%$ at $K=2,4,8$ (\Cref{fig:compute}a), while no-commitment episodes fall from 45 ($4.5\%$) to 16, 14, and 4 ($0.4\%$). The control, which launches five and retains three, records one no-commitment episode and exceeds $K=1$ by $3.3$ points. The adjacent gains shrink with $K$, so the sweep characterizes a performance--inference tradeoff (\Cref{tab:arms}) rather than fixing an optimal $K$.

For movement sensitivity, both budgets use the same $5{,}300$ replayable full-split episodes, excluding the 11 preflight aborts. SR is $44.0\%$ at 100 actions and $51.9\%$ at 200 (\Cref{fig:compute}c), a $7.9$-point difference.

\paragraph{Trajectory diagnostic.}
Across the full split, $68.6\%$ of trajectories enter the 20\,m target region and $51.8\%$ finish there; $16.8\%$ enter and then finish outside it, and $31.4\%$ never enter (\Cref{fig:failures}). Among the 890 enter-and-leave episodes, 675 complete the final transit to a selected point outside the target region, 125 end motion before a later fallback selection, 29 exhaust the budget before the final transit and 22 during it, and 39 select a point inside the region but finish outside after discrete motion. Incorrect positional selection therefore accounts for $75.8\%$ of these outcomes, and movement-budget exhaustion for $5.7\%$.

\begin{table}[t]
\centering
\caption{\textbf{Success beside recorded inference cost} on the shared $1{,}000$ episodes at 200 actions. Input tokens and generated tokens (including thinking) are per-episode means in thousands; model requests are per-episode medians. Worker hours sum episode wall clocks under parallel batch serving.}
\label{tab:arms}
\footnotesize
\setlength{\tabcolsep}{2.2pt}
\begin{tabular}{lrrrrr}
\toprule
Arm & SR & \shortstack{Input\\tokens (k)} & \shortstack{Generated\\tokens (k)} & \shortstack{Model\\requests} & \shortstack{Worker\\hours} \\
\midrule
Control & 50.0 & 1{,}639.7 & 132.6 & 115 & 372 \\
$K=1$ & 46.7 & 587.5 & 50.0 & 45 & 355 \\
$K=2$ & 49.5 & 979.7 & 78.5 & 72 & 424 \\
$K=4$ & 51.4 & 1{,}687.2 & 131.5 & 118 & 503 \\
$K=8$ & 51.8 & 2{,}908.1 & 230.8 & 196 & 611 \\
No Examiner & 44.1 & 1{,}102.5 & 102.9 & 67 & 145 \\
ViSA-inspired & 27.9 & 5.7 & 22.3 & 2 & 149 \\
GeoNav (adapted) & 27.7 & 33.6 & 42.3 & 13 & 266 \\
\bottomrule
\end{tabular}
\end{table}

\paragraph{Compute and operating points.}
\Cref{tab:arms} places each arm's success beside its recorded inference cost, and \Cref{fig:compute}d plots success against generated tokens, including thinking, for every completed reasoning-effort setting on the shared episodes. The effort setting governs the Reasoner, Examiner, and Judge calls; context compaction stays at 0.99. The adapted policy implementations make 2 (ViSA-inspired) and 13 (GeoNav) median model requests per episode, the control 115. Two contrasts summarize the sweep. At effort 0.9, WORLDS achieves $46.5\%$ SR with 38.3M generated tokens (38k per episode), against $27.7\%$ SR and 42.3M tokens (42k per episode) for adapted GeoNav at effort 0.99 on the same $1{,}000$ episodes: $18.8$ points higher at $9.4\%$ fewer generated tokens. Within WORLDS, lowering the effort from 0.99 to 0.9 cuts generated tokens by $71.1\%$ for a $3.5$-point decrease in SR. The advantage over the adapted baselines therefore does not depend on a larger generated-token budget, and the effort setting offers a measured tradeoff between these two WORLDS operating points; below 0.9, both expenditure and success are nonmonotonic in effort, and the lowest settings lack token records for episodes that aborted before flight. Over the full split, the run records a median 1.63M prompt tokens and 114 requests per episode across $5{,}300$ episodes with usage data, and 1{,}936 worker hours summed over concurrent episode wall clocks.

\subsection{Field case study on a quadrotor}
\label{sec:exp-outdoor}

\paragraph{Protocol.}
The case study asks whether the sensing waypoints WORLDS generates can be flown by a real aircraft and whether its grounding holds on the imagery that aircraft records. The aircraft is a PX4 quadrotor with GNSS localization, a Raspberry Pi 5, and an IMX290 camera. From an OSM-seeded map, WORLDS compiled sensing plans immediately prior to takeoff for three instructions, targeting an unnamed parking lot, a pool, and a red SUV absent from the map. In one 132\,s sortie, the aircraft flew the three generated waypoints at up to 60.9\,m above ground, arriving within 2.5, 5.8, and 2.6\,m of the requested locations, before a pilot-initiated return (\Cref{fig:realworld}). An image-datalink failure prevented images from reaching the reasoning agents during flight, so recorded onboard images were processed right after landing.

\paragraph{Outcomes.}
Processing the recorded images, WORLDS selected the intended target in all three tasks.

\Q{Relational grounding (Task 1).} WORLDS grounded the unnamed parking lot through its relation to a named building nearby, using the map for the relation and the recorded image for confirmation.

\Q{Conflicting language (Task 2).} The pool was selected from the recorded view although a directional cue in the instruction contradicted it; the commitment is the best match under inconsistent evidence rather than a literal reading of the instruction.

\Q{Target absent from the prior (Task 3).} The red SUV has no counterpart in the OSM prior. Perception segmented it in the parking-deck image and added it to the graph as a runtime node, and the Judge selected that node: the search grounded a target the map could not represent.

\section{Limitations}
\label{sec:method_limitations}

\paragraph{Method limitations.}
All Reasoners read the same graph, so errors in perception, map registration, or detection association reach every hypothesis and cannot be caught by comparing them. The segmentation concepts are fixed after interpretation, so an object class no Reasoner anticipated is never detected. Confidences are heuristic rather than calibrated, and the Judge may commit before every predicate is verified, so a selected target may satisfy only part of the instruction. Top-down views cannot resolve features such as facades, and a tour fixed within a pass may still visit hypotheses the evidence has already weakened.

\paragraph{Evaluation limitations.}
The ablations measure only the Examiner and the number of Reasoners. Structuring the prior as a graph, the persistence of detected objects, and question-driven view selection were never removed individually, so the baseline comparison shows that the full system outperforms the adapted baselines but not which of these is most responsible for the gain. Each arm is a single stochastic run, so variation across repeated runs, scenes, and cities is not measured, and episodes sharing a target or similar wording are not independent. In the quadrotor deployment, onboard images were processed after landing, so in-flight replanning from perception, and its latency, remain to be demonstrated.

\section{Conclusion}
\label{sec:conclusion}

We presented WORLDS, a framework for language-guided search built on a persistent world-state graph seeded from a geospatial prior and updated only by perception. Competing interpretations of an instruction are kept alive, the observations that would separate them are gathered on one shared sensing tour, and each interpretation is verified against those observations before a target is committed. On CityNav, WORLDS reaches $51.8\%$ success over all $5{,}311$ test episodes, $15.7$ points above the best published result under an OSM-only orthographic protocol, and $50.0\%$ against $27.9\%$ for the strongest adapted baseline given the same model, prior, sensing, and movement budget. The ablations attribute $5.9$ points to observation-based verification and show that more Reasoners broaden what the system asks to see at little added flight, so the architecture converts additional inference into better search, and it still exceeds adapted GeoNav at a reduced reasoning effort with fewer generated tokens. On a quadrotor, the same system flew its generated sensing waypoints and grounded, from onboard imagery, a vehicle absent from the map. Future work will close the loop in flight, so that the tour is replanned as evidence weakens hypotheses, and calibrate the confidences on which the Judge commits, so that commitment can require every predicate of the instruction to be verified.

\section*{Acknowledgments}
The authors thank the Naval Postgraduate School High Performance Computing team for their support. OpenAI Codex and Anthropic's Claude assisted with revising the writing, generating plotting and layout code for Figs.~\ref{fig:failures}, \ref{fig:compute}, and \ref{fig:realworld} from experiment records and photographs, and checking LaTeX consistency. All claims, numbers, and experimental records were verified by the authors.

\bibliography{references}
\end{document}